\documentclass[letterpaper, 10 pt, conference]{ieeeconf}  

\IEEEoverridecommandlockouts                              

\usepackage{bm}

\usepackage{color} 
\usepackage[linesnumbered,ruled,vlined]{algorithm2e}

\usepackage{mathrsfs}
\usepackage{cite}
\usepackage{amsmath,amssymb,amsfonts}
\usepackage{algorithmic}
\usepackage{graphicx}
\usepackage{subfigure}
\usepackage{textcomp}
\usepackage{xcolor}
\usepackage{diagbox}
\usepackage{color}
\usepackage{caption}  
\usepackage{graphicx}
\usepackage{float}
\usepackage{url}

\usepackage{titlesec} 

\titleformat{\subsubsection}[block]
  {\normalfont\itshape}  
  {\arabic{subsubsection})} 
  {0.6em}                 
  {}                    

\title{\LARGE \bf
An Agile Large-Workspace Teleoperation Interface Based on \\ Human Arm Motion and Force Estimation
}

\author{Jianhang Jia$^{1}$, Hao Zhou$^{1}$, and Xin Zhang$^{1,2*}$
\thanks{This work was supported by the National Natural Science Foundation of China (Grant No. 62103407).}
\thanks{$^{1}$Shenyang Institute of Automation, Chinese Academy of Sciences, Shenyang, China. $^{2}$School of Computing, University of Portsmouth, Portsmouth, United Kingdom. Email addresses: \tt\small{jiajianhang@sia.cn, zhouhao1@sia.cn,  zhangxin3@ieee.org}}%
\thanks{*Corresponding Author}%
}

\begin{document}

\maketitle
\thispagestyle{empty}
\pagestyle{empty}

\begin{abstract}

Teleoperation can transfer human perception and cognition to a slave robot to cope with some complex tasks, in which the agility and flexibility of the interface play an important role in mapping human intention to the robot.  In this paper, we developed an agile large-workspace teleoperation interface by estimating human arm behavior. Using the wearable sensor, namely the inertial measurement unit and surface electromyography armband, we can capture the human arm motion and force information, thereby intuitively controlling the manipulation of the robot. The control principle of our wearable interface includes two parts: (1) the arm incremental kinematics and (2) the grasping recognition. Moreover, we developed a teleoperation framework with a time synchronization mechanism for the real-time application.  We conducted experimental comparisons with a versatile haptic device (Omega 7) to verify the effectiveness of our interface and framework. Seven subjects are invited to complete three different tasks: free motion, handover, and pick-and-place action (each task ten times), and the total number of tests is 420. Objectively, we used the task completion time and success rate to compare the performance of the two interfaces quantitatively. In addition, to quantify the operator experience, we used the NASA Task Load Index to assess their subjective feelings. The results showed that the proposed interface achieved a competitive performance with a better operating experience. 

\end{abstract}


\section{INTRODUCTION}

Teleoperation is one of the most classic forms of human-robot interaction (HRI) \cite{sheridan2016human}, endowing human intelligence to robots. It is widely used across various fields, such as space on-orbit services \cite{zhang2020effective}, underwater exploration \cite{phung2023enhancing}, medical surgery \cite{patel2022haptic}, and other hazardous environments\cite{siciliano2008springer}. According to the human-robot interfaces, teleoperation can generally be divided into contact and contactless manners \cite{mukherjee2022survey}. For the contact manner, operators usually manipulate joysticks or haptic devices to finish the task. Although haptic devices can provide high precision control and excellent mapping relationships, their limited degrees of freedom (DoFs), workspace, and unnatural arm motion sometimes cannot meet the ergonomic requirements for human use. One clear trend is that their physical size (e.g., Virtuose 6D series from Haption company \cite{haption2024}) is made bigger to guarantee comfortable operability. In contrast, the contactless manner without manipulating limited mechanical structures is more flexible, and it captures human-centered modal information (e.g., voice, gesture, and body motion ) as input to control robots \cite{gao2024human}.
\begin{figure}[t]
\centering{\includegraphics[width=0.43\textwidth]{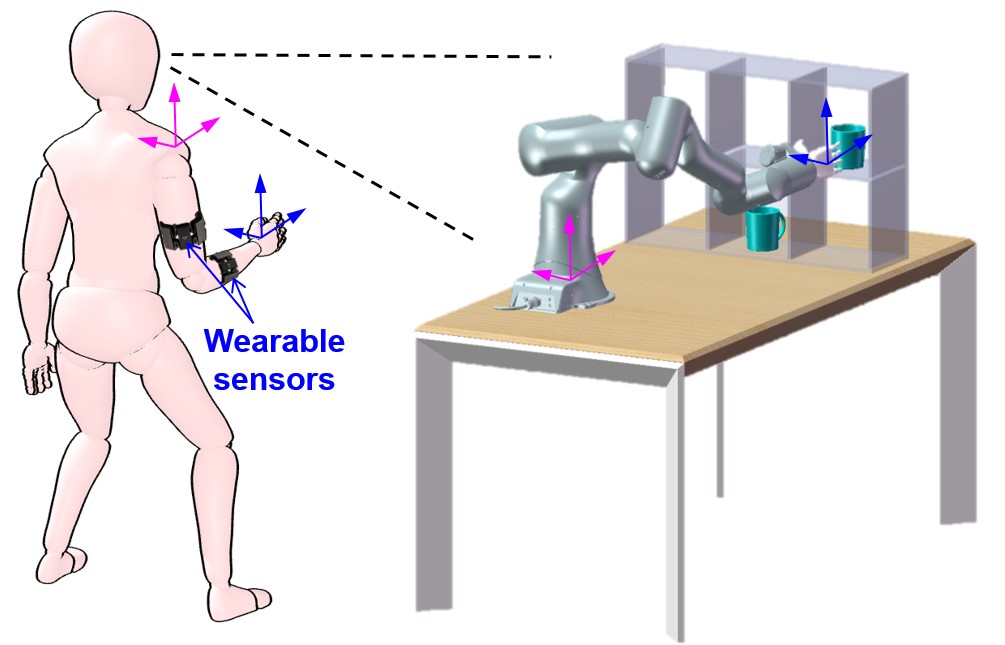}}
\caption{Concept Illustration of Cartesian Teleoperation with Two Wearable Sensors.}
\label{fig:conception}
\end{figure}

Motion capture (MoCap) technologies via cameras or wearable sensors have been developed to achieve teleoperation over the past decades.  For camera-based teleoperation, Gao et al.\cite{gao2023parallel} proposed a bimanual manipulation framework using RGB-D camera MoCap, where a parallel dual-hand detection network is designed to distinguish left and right hands. Ajoudani et al. \cite{ajoudani2012tele} proposed a teleoperation interface with impedance regulation using an optical tracking system and surface electromyography (sEMG) sensors. Bimbo et al. \cite{bimbo2017teleoperation} proposed a bilateral teleoperation interface by combining two wearable vibrotactile armbands and an RGB-D camera. Although the camera-based MoCap can provide accurate human motion feedback, the limited activity space and the multi-camera configuration are non-negligible issues. 

For the wearable MoCap, inertial measurement unit (IMU) sensors are widely used to capture human joint and whole-body motion. In our previous work \cite{du2023bi, zhang2023human}, we adopted the Xsens MVN system to capture human motion, where 17 IMU wireless modules were used to build a sensory network. In practice, there is no need to capture the human whole-body motion information because it is redundant to control a manipulator with 6 or 7 DoFs. Xu et al. \cite{xu2016development} developed a MoCap system using two MYO armbands (with IMU and sEMG data feedback) to construct the joint-space mapping of the 6-DoF arm and conducted hardware-in-loop simulations to test the proposed method. Similarly, Tortora et al.\cite{tortora2019dual} simplified the human arm with 5 DoFs and used two MYO armbands to fully capture the arm motion, where IMU data is fused to reconstruct shoulder and elbow joint motion, and sEMG data is used to estimate wrist joint motion. Kim et al. \cite{kim2017arm} proposed an algorithm to estimate the joint angle of the whole arm in 2D space with one MYO armband on the upper arm.

Unlike the joint-space teleoperation schemes \cite{xu2016development,tortora2019dual,kim2017arm}, we proposed a Cartesian teleoperation interface with two wearable armbands as shown in Fig. \ref{fig:conception}. The main contributions of this work can be summarized as:

\begin{itemize}

\item An agile large-space teleoperation interface is developed with two MYO armbands, where the Cartesian motion and the grasping action are achieved by estimating human arm motion and force. 
\item Based on our interface, a teleoperation framework is proposed to guarantee real-time synchronization mapping between the human and robotic arms.
\end{itemize}

The rest of the paper is organized as follows: Section II introduces the basic control principles of the proposed interface, including arm incremental kinematics and grasping recognition based on IMU and sEMG signals, respectively. Section III introduces the teleoperation framework, which guarantees real-time synchronization mapping between the human and robotic arms. In Section IV, three task scenarios with various metrics are designed to test and compare the proposed interface with the classic haptic device (Omega 7). Section V concludes the paper.

\section{CONTROL PRINCIPLES}
In this Section, we will introduce the basic control principles of Cartesian teleoperation. As shown in Fig. \ref{fig:principle}, Our interface consists of two armbands worn on the human arm, which 
employs the IMU signals from both armbands and the sEMG signals only from the forearm armband. Two general preconditions are given as follows:
\begin{enumerate}
    \item [{(P1)}] The inertial and shoulder frames $\Sigma _{I}$ and $\Sigma _{S}$ are set to keep the same orientation as the manipulator base frame $\Sigma _{MB}$. 
    \item [{(P2)}] The armbands can be firmly attached to the upper arm and the forearm and obtain their orientation (quaternion) with respect to (w.r.t.) $\Sigma _{I}$ steadily. 
    
\end{enumerate}

\subsection{Incremental Kinematics of Arm Motion}

\begin{figure}[t]
\centering{\includegraphics[width=0.47\textwidth]{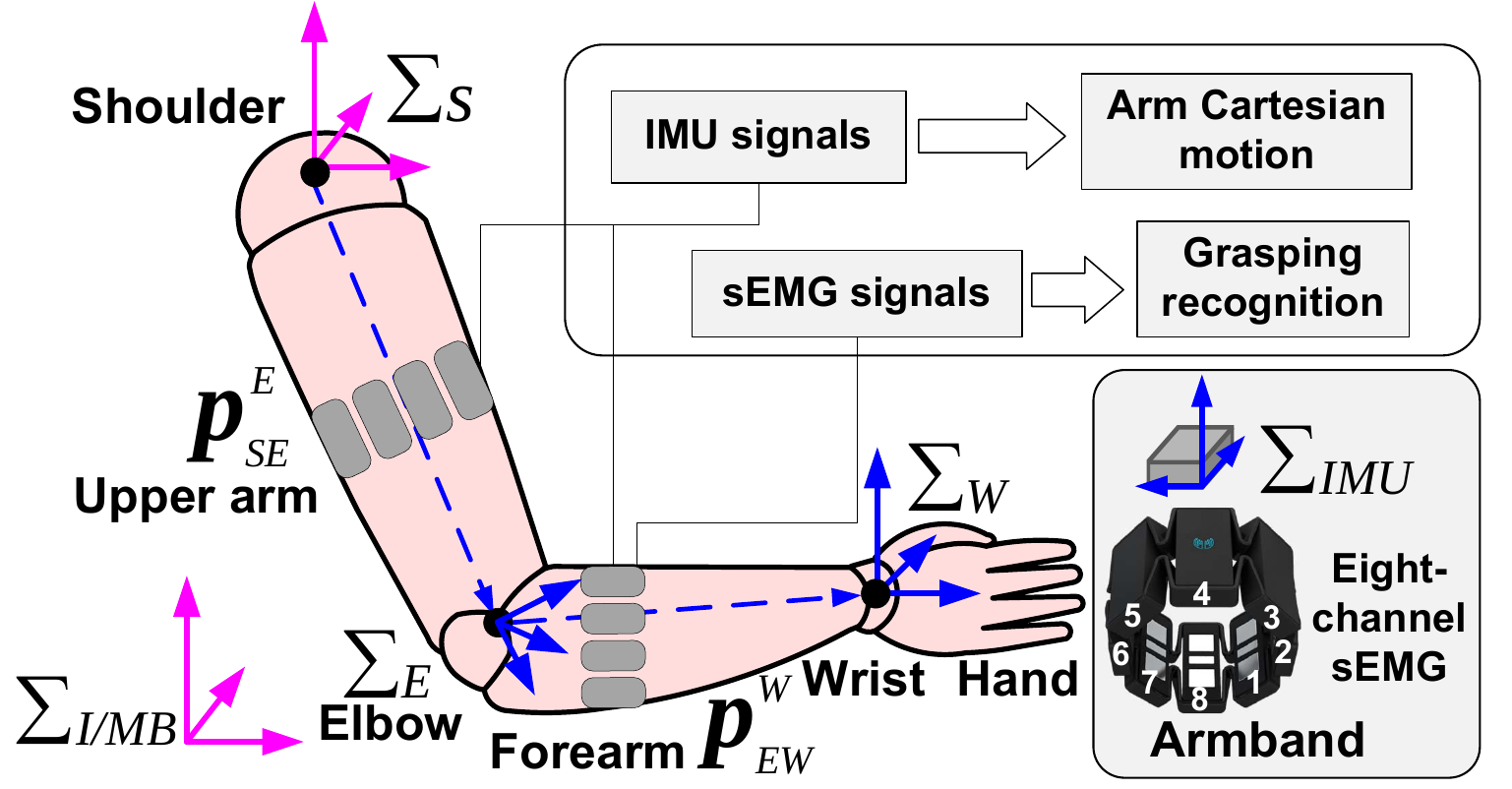}}
\caption{Principle of Human-arm Cartesian Teleoperation.}
\label{fig:principle}
\end{figure}

For Cartesian teleoperation, we employ the incremental kinematics of the arm motion to control the robotic end-effector (EE). As shown in Fig. \ref{fig:principle}, the wrist position and the forearm orientation are used to control the EE position and orientation, respectively, where $\Sigma _{MB}$ is the manipulator base frame, $\Sigma _{S}$ is the shoulder frame, $\Sigma _{E}$ is the elbow frame, $\Sigma _{w}$ is the wrist frame, $\boldsymbol{p}_{SE}^{E}=\left [ l_{U}, 0, 0 \right ]^T\in \mathbb{R}^{3}$ is the position vector from $\Sigma _{S}$ to $\Sigma _{E}$ in $\Sigma _{E}$, $\boldsymbol{p}_{EW}^{W}=\left [ l_{F}, 0, 0 \right ]^T\in \mathbb{R}^{3}$ is the position vector from $\Sigma _{E}$ to $\Sigma _{W}$ in $\Sigma _{W}$, $l_{U}$ is the length of the upper arm, and $l_{F}$ is the length of the forearm.

As we know, the IMU provides the quaternion feedback $\boldsymbol{Q}=\left \{ \eta, \boldsymbol{\epsilon} \right \}$, where $\eta$ is the scalar part and $\boldsymbol{\epsilon}=\left [ \epsilon_{x}, \epsilon_{y},\epsilon_{z} \right ]^T\in \mathbb{R}^{3}$ is the vector part. The transformation relation between the rotation matrix and the quaternion is formulated as:
\begin{equation}
\small{ \boldsymbol{R}=\begin{bmatrix}
1-2(\epsilon_{y}^2+\epsilon_{z}^2) & 2(\epsilon_{x}\epsilon_{y}-\epsilon_{z}\eta) & 2(\epsilon_{x}\epsilon_{z}+\epsilon_{y}\eta)\\ 
 2(\epsilon_{x}\epsilon_{y}+\epsilon_{z}\eta)& 1-2(\epsilon_{x}^2+\epsilon_{z}^2) & 2(\epsilon_{y}\epsilon_{z}-\epsilon_{x}\eta)\\ 
 2(\epsilon_{x}\epsilon_{z}-\epsilon_{y}\eta)&2(\epsilon_{y}\epsilon_{z}+\epsilon_{x}\eta)  & 1-2(\epsilon_{x}^2+\epsilon_{y}^2)
\end{bmatrix}}
\label{eq:transformation}
\end{equation}

Moreover, we can deduce the wrist position as
\begin{equation}
\boldsymbol{p}_{SW}^{S}=\boldsymbol{R}_{E}^{S}\cdot\boldsymbol{p}_{SE}^{E}+\boldsymbol{R}_{W}^{S}\cdot\boldsymbol{p}_{EW}^{W}
\label{eq:position}
\end{equation}
where $\boldsymbol{p}_{SW}^{S}\in \mathbb{R}^{3}$ is the position vector from $\Sigma _{S}$ to $\Sigma _{W}$ in $\Sigma _{S}$, $\boldsymbol{R}_{E}^{S}\in \mathbb{R}^{3\times3}$ represents the rotation matrix of $\Sigma _{E}$ w.r.t. $\Sigma _{S}$ obtained by (\ref{eq:transformation}) with the upper arm quaternion $\boldsymbol{Q}_{U}$, and $\boldsymbol{R}_{W}^{S}\in \mathbb{R}^{3\times3}$ represents the rotation matrix of $\Sigma _{W}$ w.r.t. $\Sigma _{S}$ obtained by (\ref{eq:transformation}) with the forearm quaternion $\boldsymbol{Q}_{F}$. 

Note that we use $\boldsymbol{p}_{SW}^{S}$ and $\boldsymbol{Q}_{F}$ to describe the pose of the human arm. Therefore, we can calculate the incremental kinematics as follows:
\begin{gather}
\Delta\boldsymbol{p}_{SW}^{S}(t)=\boldsymbol{p}_{SW}^{S}(t)-\boldsymbol{p}_{SW}^{S}(t-1)\\
\label{eq:incremental_position}
\Delta\boldsymbol{Q}_{F}(t)=\boldsymbol{Q}_{F}(t)\otimes \boldsymbol{Q}_{F}^{-1}(t-1) 
\end{gather}
where $\Delta\boldsymbol{p}_{SW}^{S}(t)\in \mathbb{R}^{3}$ represents the position increment at time $t$, $\Delta\boldsymbol{Q}_{F}(t)=\left \{ \Delta\eta_{F}(t), \Delta\boldsymbol{\epsilon}_{F}(t) \right \}$ represents the orientation increment at time $t$, and $\otimes$ is the quaternion product symbol \cite{zhang2023human}. Here, we use the quaternion to represent orientation because it can avoid the gimbal lock problem. 

\subsection{Grasping Recognition}

The armband can provide eight-channel sEMG data with various feature selections, and only the forearm armband is enough to recognize the arm muscle state (i.e., contraction or relaxation). We adopted three sEMG features: mean absolute value (MAV), waveform length (WL), and root mean square (RMS). For eight channels, we can obtain a 24-dimensional feature vector. The logistic regression algorithm \cite{kleinbaum2002logistic} is used to perform binary classification for controlling the opening and closing of the gripper.

\subsubsection{Feature Extraction}
Since sEMG signals are time series data with complex nonlinearity, feature extraction is performed within a time window set. The formulas are given as follows:
\begin{gather}
F_{MAVi}=\frac{1}{N_{t}}\sum_{j=1}^{N_{t}}\mid x_{ij}\mid \\
F_{WLi}=\sum_{j=1}^{N_{t}-1}|x_{i(j+1)}-x_{ij}| \\
F_{RMSi}=\sqrt{\frac{1}{N_{t}}\sum_{j=1}^{N_{t}}(x_{ij})^{2}}
\end{gather}
where  $x_i$ is the $i$th sEMG signal value, $F_{MALi}$ is the MAV feature, $F_{WLi}$ is the ML feature, $F_{RMSi}$ is the RMS feature, and $j=1...N_t$ is the number of samples within the time window.

\subsubsection{Binary Classification}
After obtaining the sEMG feature information from eight channels, we represent them with a feature vector.
\begin{equation}
\small{ \boldsymbol{X}=\left[F_{MAV1},F_{WL1},F_{RMS1},...,F_{MAV8},F_{WL8},F_{RMS8}\right]^T}   
\end{equation}
where $\boldsymbol{X}\in\mathbb{R}^{24}$ is the 24-D feature vector. Then, we use the logistic regression algorithm \cite{hilbe2011logistic} to perform the binary classification of the feature vector as 
\begin{gather}
P(Y=1\mid\boldsymbol{X})=\frac1{1+e^{-(\boldsymbol{W}^{T}\boldsymbol{X}+b)}}\\
P(Y=0\mid\boldsymbol{X})=1-P(Y=1\mid\boldsymbol{X})
\end{gather}
where $P(Y=1\mid\boldsymbol{X})$ is the probability of the input $\boldsymbol{X}$, $Y=1$ or $Y=0$ is the probability label, $\boldsymbol{W}\in\mathbb{R}^{24}$ is the weight vector, and b is the bias. For the decision, it can be formulated as 
\begin{gather}
D(\boldsymbol{X})=\begin{cases}
	1,&	\text{if~}P(Y=1\mid\boldsymbol{X})>0.5\\
        0,&		\text{otherwise.} \\\end{cases}   
\end{gather}
where $D(\boldsymbol{X})$ is the decision that indicates the muscle contraction, resulting in the gripper closing; otherwise, it indicates the muscle relaxation state, with the gripper opening.


\section{SYSTEM IMPLEMENTATION}

\subsection{System Framework}
The teleoperation system consists of two Linux hosts, two wearable armbands, a robotic arm, a robotic gripper, and a router. The IMU data was sampled at a frequency of 50Hz, which is much lower than the operating frequency of the robot arm 1000Hz, and the sEMG data was sampled at 200Hz. The mismatch between the IMU frequency and the robot frequency poses a great challenge to synchronizing the master-slave system. To solve the issue, two hosts and a router are adopted to build a local area network system via TCP/IP communication with non-blocking mode. This setup can reduce interference between devices with different operating frequencies and improve system performance.

\begin{figure}[t]
\vspace{-0.2cm}
\centerline{\includegraphics[width=0.48\textwidth]{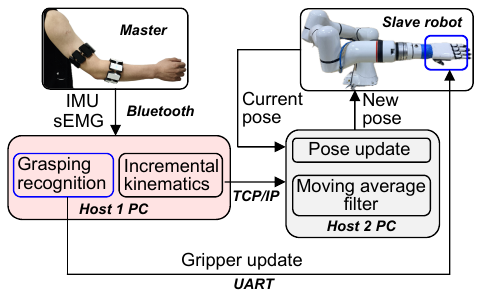}}
\caption{System Framework.} 
\label{fig:sysc} 
\end{figure}

The system framework is shown in Fig. \ref{fig:sysc}. Host 1 performs feature extraction on the sEMG data of the forearm using a time window of length $N_t$. The extracted features are formed into a feature vector $\bm{\mathit{X}}$, which is then used as input to a trained logistic regression model for binary classification. This model predicts the state of the human arm and controls the robotic gripper. Meanwhile, Host 1 uses the kinematic model (1)-(4) to calculate the position and orientation increment according to the IMU data from two armbands. 

The computed pose increment is transmitted from Host 1 to Host 2 via the local area network. 
A sliding average filter with a window size of $N$ ($N=10$) is applied to smooth the data, which can effectively reduce the impact of the human arm shaking, ensuring the stability and smoothness of the robotic arm control. The filter is formulated as follows:
\begin{equation}
\Delta\boldsymbol{p}_{avg}(t)=\frac1{N}\sum_{{i=1}}^{N}\Delta\boldsymbol{p}_{SWi}^{S}    
\end{equation}
where $\Delta\boldsymbol{p}_{avg}(t)\in\mathbb{R}^{3}$ represents the position change vector after moving average filtering at time $t$.

Quaternion data is also smoothed using a similar approach. This method involves constructing a $4\times N$ matrix $\bm{\mathit{M}}$ from N quaternions and then performing singular value decomposition. 
\begin{gather}
\bm{\mathit{
M}}=
\begin{bmatrix}
\Delta\boldsymbol{Q}_{F}\left(\scriptstyle{t-N}\right)&\Delta\boldsymbol{Q}_{F}\left(\scriptstyle{t-N+1}\right)&...&\Delta\boldsymbol{Q}_{F}\left(\scriptstyle{t}\right)
\end{bmatrix}\\
\bm{\mathit{
M}}=\bm{\mathit{U\Sigma V^T}}
\end{gather}



Moreover, we extract the first column vector $\bm{\mathit{u_1}}$ from the orthogonal matrix $\bm{\mathit{U}}$ as the principal singular vector. Normalizing $\bm{\mathit{u_1}}$ yields the average quaternion $\Delta\boldsymbol{Q}_{avg}(t)\in\mathbb{R}^{4}$ at time $t$.
\begin{equation}
\Delta\boldsymbol{Q}_{avg}(t)=\frac{\bm{\mathit{u_1}}}{\left\|\bm{\mathit{u_1}}\right\|}    
\end{equation}
The position and quaternion increments after the sliding average filter are used to update parameters to control the robot position vector $\boldsymbol{p}_{R}$ and quaternion $\boldsymbol{Q}_{R}$ from time $t$ to $t+1$. 
\begin{gather}
\label{eq:16}
\boldsymbol{p}_{R}(t+1) = \boldsymbol{p}_{R}(t) + \Delta\boldsymbol{p}_{avg}(t)\\
\label{eq:17}
\boldsymbol{Q}_{R}(t+1) = \Delta\boldsymbol{Q}_{avg}(t) \otimes  \boldsymbol{Q}_{R}(t) 
\end{gather}

By iteratively updating the motion data, the desired pose of the robotic arm for teleoperation can be achieved.

\subsection{Motion Mapping Synchronization}

\begin{algorithm}[h]
    \small
    \caption{Motion Mapping Synchronization of Teleoperation}
    \label{alg:host_communication}
     \KwIn{IMU $\boldsymbol{Q}_{U}$, $ \boldsymbol{Q}_{F}$ and sEMG ($ x_i, i=1...8$)}
    \KwOut{Robot pose $\boldsymbol{p}_R, \boldsymbol{Q}_R$ and gripper state $D$}
    \BlankLine

    \SetKwBlock{HostOne}{Host1}{}
    \SetKwBlock{HostTwo}{Host2}{}
    
    \HostOne{
        \tcp{200Hz}
            \While{\textnormal{Thread\_sEMG}}
            {
                \For{$j \gets 1$ \textbf{to} $100$}{
                    $x_{ij}$ $\gets$ \textbf{ReadsEMG}()\;
                }
                $F_{MAVi}, F_{WLi}, F_{RMSi}$ $\gets$ \textbf{Features}($x_{ij}$)\;
                $\boldsymbol{X}$ $\gets$ $F_{MAVi}, F_{WLi}, F_{RMSi}$\;
                $D(\boldsymbol{X})$ $\gets$ \textbf{Logistic\_Regression}($\boldsymbol{X}$)\;
                
            }
        \BlankLine
        \tcp{50Hz}
        \While{\textnormal{Thread\_IMU}}
        {
            {
                $\boldsymbol{R}_{E}^{S}(t), \boldsymbol{R}_{w}^{S}(t)$ $\gets$ \textbf{ReadIMU}()\; 
                \begingroup                \scriptsize{$\Delta\boldsymbol{p}_{SW}^{S}(t), \Delta\boldsymbol{Q}_{F}(t)$ $\gets$ \textbf{Kinematic\_Model}($\boldsymbol{R}_{E}^{S}(t), \boldsymbol{R}_{w}^{S}(t)$)}\;
                \endgroup
                \textit{message} $\gets$ \textbf{Create\_Message}(\textit{$\Delta\boldsymbol{p}_{SW}^{S}(t), \Delta\boldsymbol{Q}_{F}(t)$})\;
                \textbf{Send\_Message}(\textit{message}, \textit{Host2\_address})\;
            }
        }
    }
    \BlankLine
    \HostTwo{
        \tcp{250Hz}
        \While{\textnormal{Thread\_Receive}}
        {
           {                              \If{\textbf{Check\_Message}(\textit{Host1\_address})}{
                    \textit{message} $\gets$ \textbf{Receive\_Message}(\textit{Host1\_address})\;
                    \begingroup                    \scriptsize{$\Delta\boldsymbol{p}_{SW}^{S}(t), \Delta\boldsymbol{Q}_{F}(t)$ $\gets$ \textbf{Extract\_Data}(\textit{message})}\;
                    \endgroup
                }
            }
        }
        \BlankLine
        \tcp{1000Hz}
        \While{\textnormal{Thread\_Control}}
        {
            {
                \begingroup                \scriptsize{$\Delta\boldsymbol{p}_{avg}(t), \Delta\boldsymbol{Q}_{avg}(t)$ $\gets$ \textbf{Smooth}($\Delta\boldsymbol{p}_{SW}^{S}(t), \Delta\boldsymbol{Q}_{F}(t)$)}\;
                \endgroup
                \begingroup             \scriptsize{$\boldsymbol{p}_{R}(t+1), \boldsymbol{Q}_{R}(t+1)$ $\gets$ \textbf{Pose}($\Delta\boldsymbol{p}_{avg}(t), \Delta\boldsymbol{Q}_{avg}(t)$)}\;
                \endgroup
                \textbf{Control\_Robot}($\boldsymbol{p}_{R}(t+1), \boldsymbol{Q}_{R}(t+1)$)\;
            }
        }
    }
\end{algorithm}

Host 2 is to coordinate the working frequencies between two armbands and the robotic arm. The frequency of armbands determines that Host 1 sends data at 50Hz, so the data update frequency for Host 2 in the motion control process is also 50Hz. However, due to the instability of  Bluetooth transmission of armbands, we incorporate a 250Hz data-checking mechanism into Host 2 to check if Host 1 has sent the position and orientation increments early or late. If Host 1 sends the data early, Host 2 updates the data accordingly. If it is delayed, Host 2 continues to use the previous data to control the robotic arm. The motion mapping synchronization is provided in Algorithm \ref{alg:host_communication}.

\begin{figure}[t]
\centering{\includegraphics[width=0.48\textwidth]{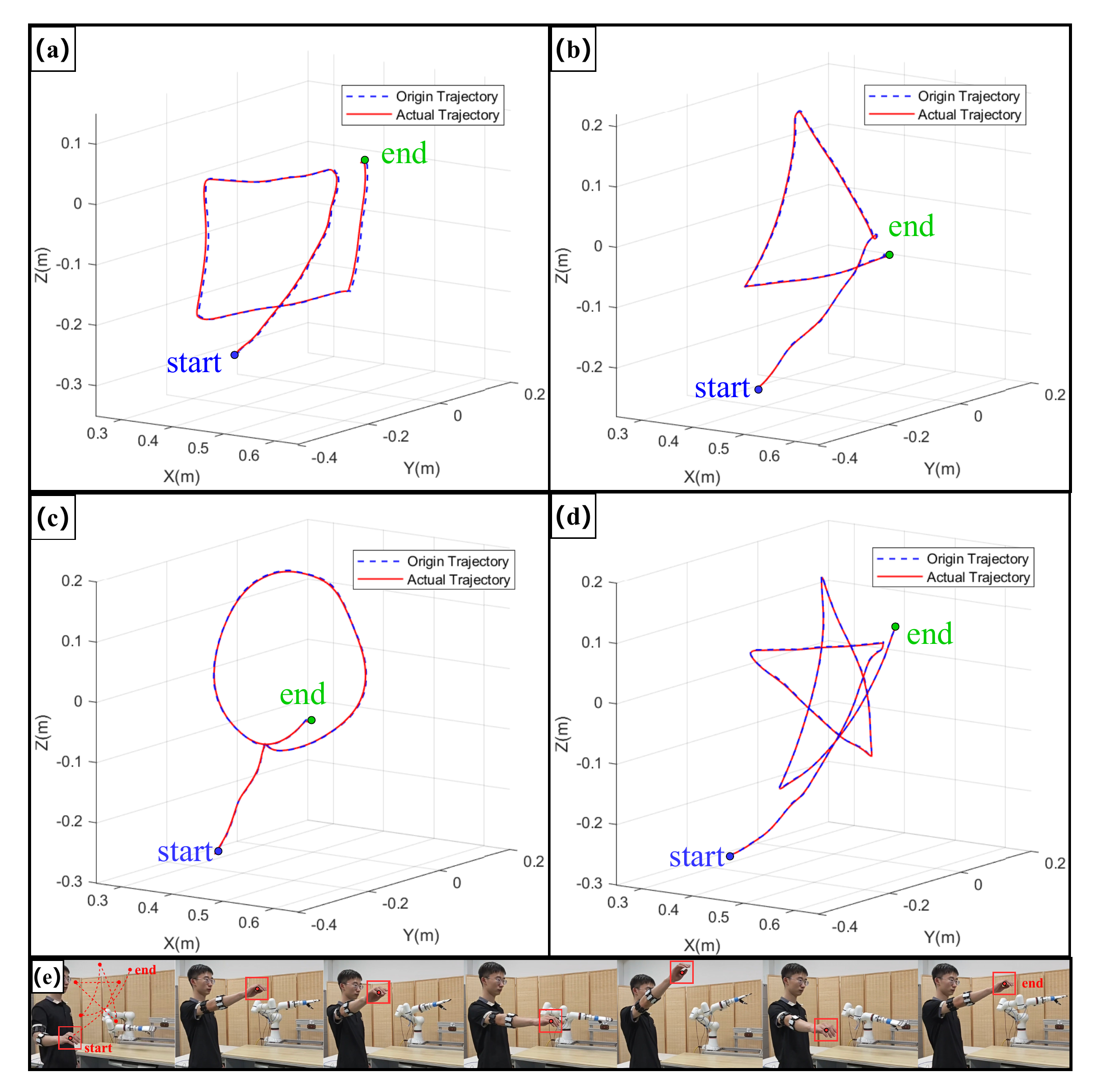}}
\caption{Motion Mapping Synchronization Experiment. (a) Square. (b) Triangle. (c) Circle. (d) Pentagram. (e) Pentagram Motion Demonstration}
\label{fig:moma}
\end{figure}

\begin{table}[t]
\centering 
 TABLE \uppercase\expandafter{\romannumeral 1}. Evaluation of Synchronization\vspace{0.16cm}
\renewcommand{\arraystretch}{1.5}
\resizebox{0.47\textwidth}{!}{%
\begin{tabular}{ccccccccc}
\hline
           & \multicolumn{2}{c}{\textbf{Square}} & \multicolumn{2}{c}{\textbf{Triangle}} & \multicolumn{2}{c}{\textbf{Circle}} & \multicolumn{2}{c}{\textbf{Pentagram}} \\
           & \textbf{S\textsubscript{RMSE}}     & \textbf{S\textsubscript{MAE}}    & \textbf{S\textsubscript{RMSE}}      & \textbf{S\textsubscript{MAE}}     & \textbf{S\textsubscript{RMSE}}     & \textbf{S\textsubscript{MAE}}    & \textbf{S\textsubscript{RMSE}}      & \textbf{S\textsubscript{MAE}}      \\ \hline
\textbf{X(m)} & 0.0080            & 0.0060          & 0.0073             & 0.0054           & 0.0060            & 0.0044          & 0.0062             & 0.0046            \\
\textbf{Y(m)} & 0.0124            & 0.0076          & 0.0126             & 0.0078           & 0.0135            & 0.0086          & 0.0122             & 0.0091            \\
\textbf{Z(m)} & 0.0143            & 0.0091          & 0.0150             & 0.0106           & 0.0147            & 0.0099          & 0.0165             & 0.0124            \\ \hline
\end{tabular}%
}\vspace{-0.05cm} 
\end{table}

To verify the motion mapping synchronization performance, we invited several subjects to wear armbands and do some free motion, namely four shapes: square, triangle, circle, and pentagram. During these processes, we recorded the motion data of armbands and the actual motion data of the robotic arm, as shown in Fig. \ref{fig:moma}. Furthermore, we evaluated the synchronization and motion distance accuracy between our teleoperation interface and the robotic arm using root mean square error (RMSE) and mean absolute error (MAE).
\begin{gather}
S_{RMSE}=\sqrt{\frac{1}{n}\sum_{i=1}^n\left(T_{Hi}-T_{Ri}\right)^2}\\
S_{MAE}=\frac{1}{n}\sum_{i=1}^n\mid T_{Hi}-T_{Ri}\mid
\end{gather}
where $T_{Hi}$ represents the trajectory data of the human arm and $T_{Ri}$ denotes the actual trajectory data of the robotic arm.

\begin{figure}[t]
\centering{\includegraphics[width=0.48\textwidth]{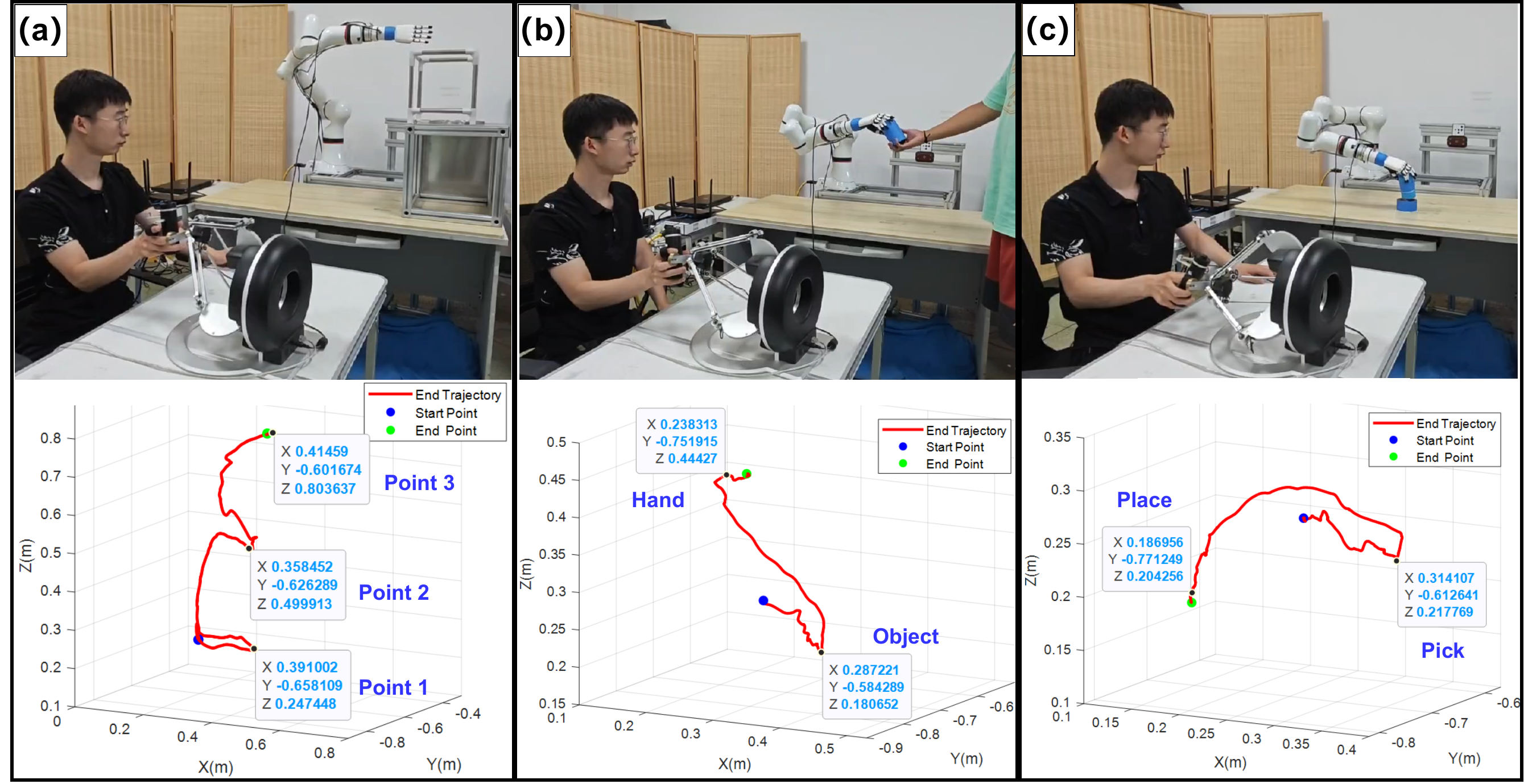}}
\caption{Traditional Interface. (a) Free Motion. (b) Handover. (c) Pick and Place.} 
\label{fig:joy} 
\end{figure}
Table \uppercase\expandafter{\romannumeral 1} shows the evaluation metric values for the spatial positions between the human arm trajectory and the robotic arm trajectory after completing the four shapes using both interfaces. X, Y, and Z represent the three coordinate directions in $\Sigma _{MB}$, respectively. The $S_{RMSE}$ is less than $0.0165~m$, and the $S_{MAE}$ is less than $0.0124~m$. This indicates a high tracking accuracy between the human and robotic arms and demonstrates the feasibility of the proposed framework using wearable armbands.

\begin{figure*}[t]
\centerline{\includegraphics[width=0.95\textwidth]{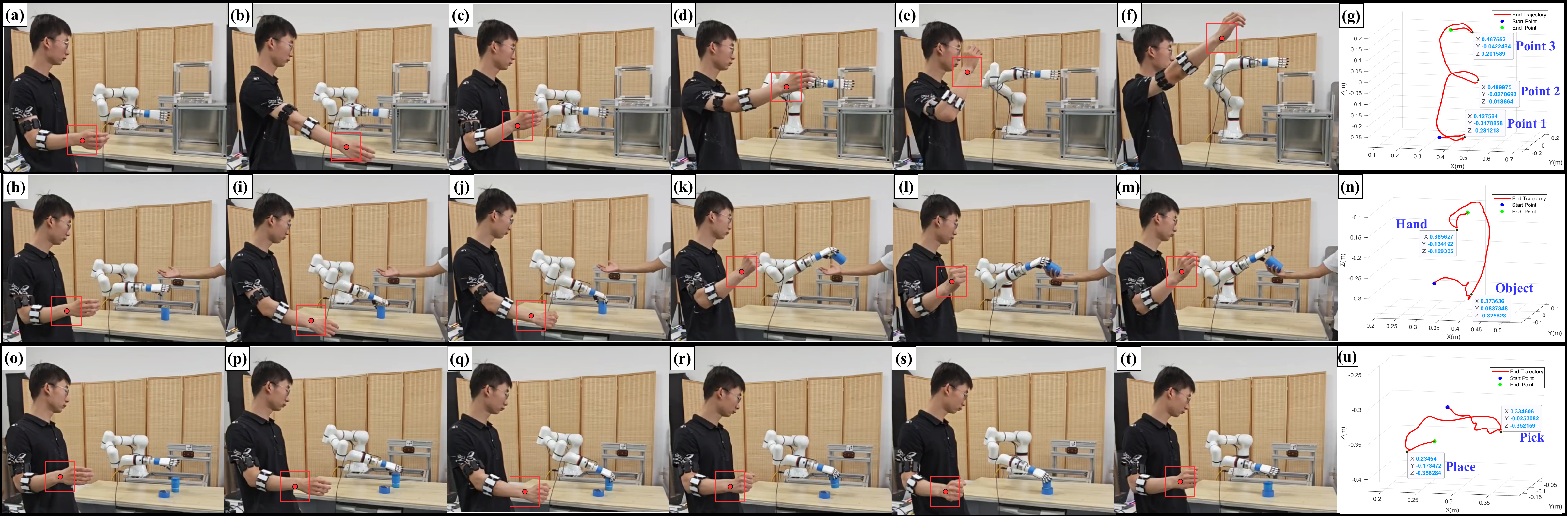}}
\caption{Wearable Interface.  (a-g) Free Motion. (h-n) Handover. (o-u) Pick and Place.} 
\label{fig:wear}  \vspace{-0.21cm} 
\end{figure*}

\section{COMPARATIVE STUDIES}

 To validate the natural interactivity and effectiveness of our interface in large-workspace teleoperation tasks, we invited seven subjects aged between 24 and 28 years old to test our interface and the traditional interface (Omega 7) to perform various tasks in a comparative study. The three experimental tasks included Free Motion, Handover, and Pick and Place. Each subject completed ten operations per task with each interface, 420 tests in total. Before the experiments, we provided each subject with an overview of the key considerations, initial pose correction, and preliminary interface manipulation to ensure their basic understanding and familiarity. The specific operational instructions for the three experimental tasks are:

\begin{itemize}

\item Free Motion: Subjects performed obstacle-avoidance movements in 3-D space, in which the EE orientation is not controlled. 
\item Handover: Subjects manipulated the robotic arm to grasp an object on the table and hand it to a receiver. 
\item Pick and Place: Subjects manipulated the robotic arm to accurately grasp an object and place it at the designated location.

\end{itemize}

Free Motion primarily tests position teleoperation in a large workspace; handover aims to test the natural interaction experience between a human arm and a robotic arm; and the pick-and-place experiment is used to test the precision of robotic arm control.

\subsection{Traditional interface}

The haptic device Omega.7, as one of the most traditional interfaces, is used to perform experimental tasks and record the processes and the robot EE trajectory, as shown in Fig. \ref{fig:joy}. When executing large-space teleoperation tasks, the traditional device is limited by its operational space. It needs to map its small operational space to a large space of the robot arm with a 100-times amplification coefficient. This mapping from a small space to a large space can cause the robotic arm to be overly sensitive, amplifying the tremor transmission from the human arm to the robotic arm. Therefore, we can see that the recorded EE trajectory is a little uneven, but it still can complete the tasks successfully in a large space.

\begin{figure}[t]
\vspace{0.1cm} 
\centering{\includegraphics[width=0.48\textwidth]{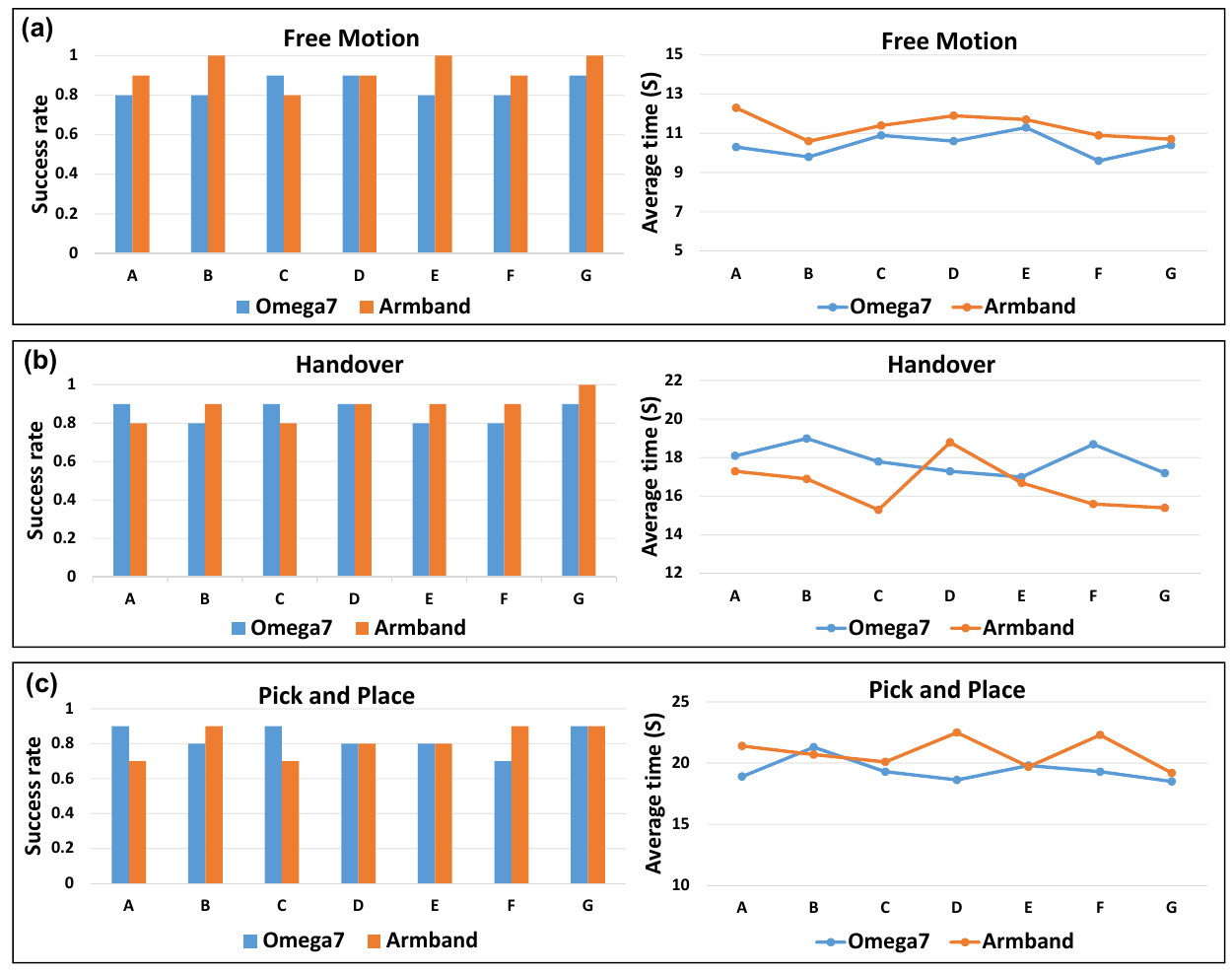}}
\caption{Comparison of Success Rates and Average Operation Times in the Three Tasks. (a) Free Motion. (b) Handover. (c) Pick and Place.}\vspace{-0.08cm}
\label{fig:suti}
\end{figure}

\begin{figure}[t]
\centering{\includegraphics[width=0.47\textwidth]{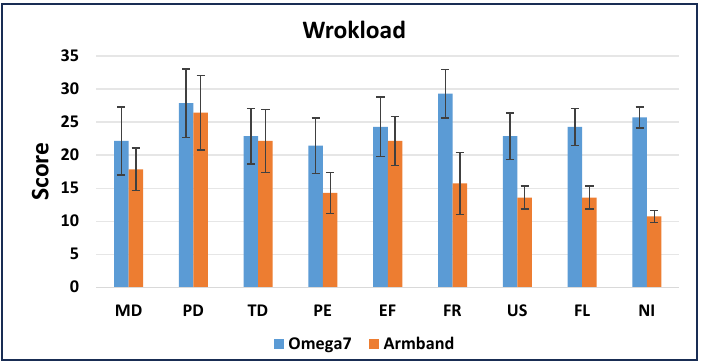}}
\caption{Extended NASA-TLX Assessment.}
\label{fig:nasa}
\end{figure}

\subsection{Wearable interface}

Our wearable devices are two armbands. The motion data of the human arm is calculated using the IMU data from the armbands. The experimental processes and the robot EE trajectory are shown in Fig. \ref{fig:wear}. Due to the high flexibility of human arm movements, the robotic arm controlled by the user exhibits strong flexibility and natural interactivity, providing relatively smooth and stable control with good synchronization. This enables efficient completion of the three experimental tasks without robotic arm tremors.

\subsection{Results and analysis}

Subjective evaluation metrics based on the NASA Task Load Index (TLX) \cite{hart2006nasa} and two objective evaluation metrics (task success rate and average completion time) were used to compare the two interfaces. The NASA-TLX includes six assessment metrics: Mental Demand (MD), Physical Demand (PD), Temporal Demand (TD), Performance (PE), Effort (EF), and Frustration (FR). To gain a more comprehensive evaluation of the NASA-TLX form, three additional metrics were included: Usability (US) assesses the ease of task execution by the system; Fluency (FL) measures the smoothness of collaboration between the robot and the subject; and Natural Interaction (NI) evaluates the system's ability to interact naturally with the user. Note that lower scores indicate better task performance in the extended NASA-TLX metrics. 

\begin{table}[t]
\centering
\vspace{0.39cm} TABLE \uppercase\expandafter{\romannumeral 2}. Summary of Average Times and Success Rates\vspace{0.16cm}
\renewcommand{\arraystretch}{1.5}
\resizebox{0.47\textwidth}{!}{%
\begin{tabular}{ccccccc}
\hline
\textbf{}                & \multicolumn{2}{c}{\textbf{Free Motion}} & \multicolumn{2}{c}{\textbf{Handover}} & \multicolumn{2}{c}{\textbf{Pick and Place}} \\
\textbf{}                & \textbf{Omega7}      & \textbf{Armband}      & \textbf{Omega7}     & \textbf{Armband}    & \textbf{Omega7}        & \textbf{Armband}       \\ \hline
\textbf{Average time(s)} & 10.4                 & 11.3              & 17.9                & 16.6            & 19.4                   & 20.8               \\
\textbf{Success rate}    & 84\%                  & 93\%               & 85\%                 & 89\%             & 83\%                    & 82\%                \\ \hline
\end{tabular}
}
\vspace{-0.05cm} 
\end{table}

Statistic results on task success rates and average completion times under different interfaces are presented in Fig. \ref{fig:suti}. The average time and success rates for all subjects across the three experimental tasks are summarized in Table \uppercase\expandafter{\romannumeral 2}. In free motion tests, the wearable interface demonstrated a higher success rate, exceeding the traditional joystick interface by 9\%. However, the completion time was slightly longer, about 1 second more than the traditional interface, because subjects moved within a larger workspace, which took more time. In handover tests, the system significantly outperformed the traditional interface regarding natural interaction, with a 1-second faster completion time and a 4\% higher success rate approximately. In pick-and-place tests, the completion rate was comparable to the traditional interface, with only a 1\% difference, but the traditional interface slightly outperformed the system in terms of completion time.

According to the extended NASA-TLX assessment results in Fig. \ref{fig:nasa}, the wearable sensor interface provides a superior overall experience in performing teleoperation tasks compared to the traditional joystick device. This is particularly evident in tasks involving large-scale teleoperation and natural human-robot interaction, where the wearable human-robot interface attributes and natural flexibility are more valued. The wearable sensor interface excels in operational smoothness and synchronization and offers a better user experience during prolonged tasks, reducing operational fatigue and frustration. These advantages make it particularly effective in complex environments and tasks, aligning well with efficient and natural interaction demands.

\section{CONCLUSION AND FUTURE WORK}

In this work, we proposed an agile large-workspace interface and a framework using two IMU-sEMG armbands for the Cartesion teleoperation. The proposed teleoperation framework estimated the human arm motion and force for controlling the slave robot. To demonstrate the advantages of the proposed method, we compared it with the traditional method via three manipulation tasks. In the free motion and handover tasks, the wearable interface showed superior performance and a more natural interaction experience than the traditional interface. The wearable interface performs competitively for tasks requiring extremely high precision (such as the pick-and-place task). The extended NASA-TLX evaluation further confirmed that the wearable interface offered a better user experience and a more natural interaction experience. In addition, the wearable interface can provide a low-cost solution for teleoperation.

However, we also met a limitation in our work, i.e., the drift issue of IMU sensors during prolonged tasks, potentially affecting the accuracy of the arm pose data. In the future, we plan to incorporate vision technology to mitigate the effects of IMU drift.

\bibliographystyle{IEEEtran}
\bibliography{ROBIO}{}

\end{document}